\documentclass[conference]{IEEEtran}
\IEEEoverridecommandlockouts
\usepackage{cite}
\usepackage{amsmath,amssymb,amsfonts}
\usepackage{algorithm}
\usepackage{algorithmic}
\usepackage{graphicx}
\usepackage{booktabs}
\graphicspath{{figures/}}
\usepackage{textcomp}
\usepackage{xcolor}
\def\BibTeX{{\rm B\kern-.05em{\sc i\kern-.025em b}\kern-.08em
    T\kern-.1667em\lower.7ex\hbox{E}\kern-.125emX}}

\usepackage{comment}
\usepackage{balance}
\usepackage{tikz}
\usetikzlibrary{arrows.meta,positioning,shapes,calc,fit}

\usepackage[svgnames]{xcolor}
\usepackage[colorlinks=true,
            breaklinks=true,
            linkcolor=DarkOrchid,
            urlcolor=cyan,
            citecolor=Maroon]{hyperref}
\usepackage[capitalize,noabbrev]{cleveref}
\newcommand{\Albireo}{\textsc{Albireo}}

\makeatletter
\newcommand{\linebreakand}{%
  \end{@IEEEauthorhalign}\hfill\mbox{}\par
  \mbox{}\hfill\begin{@IEEEauthorhalign}}
\makeatother

\begin{document}

\title{\Albireo: Adaptive, Energy-Efficient  Inference Framework for Video Object Detection on the Edge}

\author{%
\IEEEauthorblockN{Amir Taherin}
\IEEEauthorblockA{\textit{Northeastern University}\\
Boston, MA, USA\\
taherin.a@northeastern.edu}
\and
\IEEEauthorblockN{Jos\'e Cano}
\IEEEauthorblockA{\textit{University of Glasgow}\\
Glasgow, UK\\
Jose.CanoReyes@glasgow.ac.uk}
\and
\IEEEauthorblockN{Bin Ren}
\IEEEauthorblockA{\textit{William \& Mary}\\
Williamsburg, VA, USA\\
bren@cs.wm.edu}
\linebreakand
\IEEEauthorblockN{Yanzhi Wang}
\IEEEauthorblockA{\textit{Northeastern University}\\
Boston, MA, USA\\
yanzhiwang@northeastern.edu}
\and
\IEEEauthorblockN{David Kaeli}
\IEEEauthorblockA{\textit{Northeastern University}\\
Boston, MA, USA\\
d.kaeli@northeastern.edu}
}

\maketitle
\begingroup\renewcommand\thefootnote{}\footnotetext{\copyright~2026 IEEE.
Personal use of this material is permitted. Permission from IEEE must be
obtained for all other uses, in any current or future media, including
reprinting/republishing this material for advertising or promotional
purposes, creating new collective works, for resale or redistribution to
servers or lists, or reuse of any copyrighted component of this work in
other works. Accepted at the ACM/IEEE Symposium on Edge Computing (SEC) 2026.}\endgroup

\begin{abstract}
Video object detection on edge devices requires running computationally expensive detectors over long frame streams, leading to high energy consumption and sustained GPU utilization. 
Although consecutive video frames contain substantial temporal redundancy, naive frame-skipping strategies are content-blind: they skip during critical moments such as object entry, occlusion recovery, and abrupt motion, causing substantial detection-quality loss. 
We present \Albireo{}, a detector-agnostic, codec-free, adaptive inference framework that wraps off-the-shelf object detectors and decides when detector invocation can be safely skipped based on scene content and per-object temporal state. 
\Albireo{} requires no detector modification or detector retraining and is designed as a drop-in efficiency layer for improving the accuracy--energy--latency tradeoff of edge video-detection pipelines.

To judge when skipping is safe, \Albireo{} maintains a 10-dimensional Kalman filter (KF) for each active object state and invokes the detector only when prediction uncertainty exceeds a threshold. 
On skipped frames, bounding boxes are predicted from the KF state at near-zero GPU cost. 
A KF-based rescue mechanism preserves confirmed object states through brief detector misses, preventing output fragmentation that degrades detection-to-ground-truth matching. 
A lightweight empty-scene screen further avoids full detector calls on objectless frames. 

We evaluate \Albireo{} on the BDD100K dataset's multi-object tracking (MOT) validation split using three architecturally distinct detectors (YOLO11x, YOLO26x, and RF-DETR-Large) on two NVIDIA Jetson platforms (AGX Thor and AGX Orin). 
Across all detector--platform configurations, \Albireo{} keeps AP@50 (average precision at IoU 0.5) within $\pm1.2$\,pp of per-frame inference while reducing total energy by $12.1$--$17.6\%$. 
On the primary YOLO26x configuration, \Albireo{} improves AP@50 by $+0.8$\,pp while reducing energy by $17.6\%$ on Thor and $14.4\%$ on Orin. 
It also reduces per-frame energy-delay product by $24.9\%$ on Thor and $26.1\%$ on Orin, showing that the default operating point improves accuracy, energy, and latency together rather than trading one for another. 
In contrast, FixedSkip-2, a fixed-interval baseline with $50\%$ skip rate, loses $8.6$\,pp AP@50 ($14\%$ relative). \Albireo{}'s source code, evaluation pipeline, and per-clip result data are available open-source at \url{https://github.com/amirtaherin/albireo}.
\end{abstract}


\section{Introduction}
\label{sec:introduction}

Edge video object detection has matured into a production technology~\cite{azevedo2024yolo_adas, sathyamoorthy2025ensemble, kim2024statues, contoli2024dipm}. Dashcams~\cite{azevedo2024yolo_adas, lyssenko2024cflow}, traffic-monitoring nodes~\cite{sathyamoorthy2025ensemble}, autonomous vehicles~\cite{lee2024uncertaintytrack}, and security cameras~\cite{kim2024statues} now routinely run convolutional or transformer-based detectors on dedicated edge systems, such as the NVIDIA Jetson family~\cite{dutt2023evaluating, alrafi2024uncovering, holly2020profiling, taherin2026glsvlsi}. 
The dominant cost in these pipelines is the per-frame forward pass through the detector. Based on our empirical study on the BDD100K dataset's multi-object tracking (MOT) validation split (200 dashcam clips, 7 object classes)~\cite{yu2020bdd100k}, running YOLO26x~\cite{sapkota2025yolo26} on every frame draws $70$\,W of total board power on Jetson AGX Thor~\cite{nvidia_thor_trm_2025} (Blackwell, $130$\,W TDP), of which $34$\,W ($49\%$) is the GPU power; on the AGX Orin~\cite{nvidia_orin_manual} (Ampere, $60$\,W TDP), the same workload draws $43$\,W total of which $31$\,W ($72\%$) is the GPU power. 
As edge deployments expand and battery and thermally-constrained devices~\cite{sun2020arvr} become common in emerging technologies~\cite{lin2026vote, rupprecht2026survey}, reducing this per-frame cost is no longer just an optimization problem; it directly determines which perception workloads are deployable in practice.

An obvious mitigation approach is to skip frames. Video possesses a high degree of temporal redundancy~\cite{bobick2001temporal, chen2018scaletime, habibian2021skipconv, liu2018mobile}, and a well-chosen subset of frames should suffice. The obvious mitigation, applied naively, fails. Fixed-cadence skipping (the simplest content-blind baseline studied in prior work~\cite{nishimura2022sdof, park2020frameskip, henning2023framedrop}) drops accuracy sharply: on the BDD100K MOT~\cite{yu2020bdd100k}, skipping every second frame loses $8.6$\,pp~AP@50 (average precision at IoU 0.5; $14\%$ relative), and skipping four out of every five frames loses $28.8$\,pp ($47\%$). 
A fixed schedule cannot detect when a new object enters the frame, when an occlusion clears, or when an object accelerates. The design problem is not whether to skip, but instead \emph{which} frames to skip.  This decision must be informed by scene content and object-level spatial and temporal state.

Adaptive alternatives exist~\cite{zhu2017dff, shi2024yolov, kim2024statues, ding2024ctd, wang2019compressed, wang2025mayolo, luo2019detect, arefeen2022framehopper, henning2023framedrop, contoli2024dipm}, but no existing approach simultaneously provides the properties needed for drop-in edge deployment: detector-agnostic operation, no detector retraining, codec-free execution, ego-motion robustness, and demonstrated energy efficiency on commercial edge SoCs. 
Temporal feature-propagation methods~\cite{zhu2017dff, zhu2017fgfa, shi2024yolov} couple the savings to a detector backbone through architecture-specific temporal modules or retraining. 
Pixel-level frame-differencing schemes~\cite{kim2024statues} assume spatially aligned backgrounds and become unreliable under ego-motion. 
Compressed-domain methods~\cite{wang2019compressed, wang2025mayolo, duche2026compressed, yao2026comprivdet} depend on H.264/H.265 motion vectors and residuals that are unavailable in raw camera feeds before encoding. 
Confidence-Triggered Detection (CTD)~\cite{ding2024ctd} is the closest method mechanistically, but compares the Kalman prediction against the most recent detector output, so the reference becomes stale across consecutive skipped frames. 
These limitations leave a gap for a detector-agnostic, codec-free frame-skipping system that is not tied to fixed-camera assumptions and improves the accuracy--energy--latency tradeoff on commercial edge SoCs.

We present \Albireo{}\footnote{The name refers to Albireo, a visually contrasting binary star system. We use the brighter component as an analogy for full detector inference and the dimmer companion for lightweight Kalman prediction.}, a frame-skipping system designed against this gap.
\Albireo{} is, to our knowledge, the first detector-agnostic, codec-free, content-aware frame-skipping system for edge video object detection that requires no detector modification or detector retraining, and demonstrates an improved accuracy--energy--latency tradeoff on commercial edge SoCs.
Three mechanisms compose its design:

\begin{enumerate}
    \setlength{\itemsep}{0.5em}
    \item \emph{Forward-looking uncertainty trigger}: a per-object Kalman filter monitors position, size, and their temporal derivatives in normalized coordinates; detector invocation is skipped when the projected covariance trace of every active object state remains below a threshold $u$. 
    Unlike stale-detection confidence tests, this trigger is driven by covariance growth in the current object state rather than by comparison to a previous detector output. 
    
    \item \emph{Detector-flicker rescue}: on detector frames, a confirmed object state that remains unmatched can emit its Kalman-predicted box, called a \textit{rescue,} if its most recent real detector match has high confidence and its consecutive unmatched-frame count remains below a fixed cap. This prevents brief detector misses from fragmenting object outputs. 
    
    \item \emph{Empty-scene screen}: when no active object states exist, a lightweight binary empty-scene classifier gates the full detector, skipping objectless frames without invoking the expensive detector forward pass. The empty-scene classifier is fine-tuned for the evaluation domain, but the object detector itself is unchanged.
\end{enumerate}

We evaluate \Albireo{} on the BDD100K MOT across three architecturally distinct detectors (YOLO11x, YOLO26x, RF-DETR-Large) and two Jetson SoC generations (AGX Thor~\cite{nvidia_thor_trm_2025} and AGX Orin~\cite{nvidia_orin_manual}). 
On the primary YOLO26x configuration, at its default operating point, \Albireo{} improves the accuracy--energy--latency tradeoff over Vanilla per-frame inference: on Thor, it improves AP@50 by $+0.8$\,pp while reducing energy by $17.6\%$, frame time by $8.8\%$, and per-frame energy-delay product (EDP) by $24.9\%$; on Orin, it achieves the same AP@50 gain with $14.4\%$ lower energy and $26.1\%$ lower EDP. 
Across all evaluated detectors and platforms, \Albireo{} keeps AP@50 within $1.2$\,pp of Vanilla per-frame inference while reducing energy by $12.1$--$17.6\%$. 
Sweeping the primary uncertainty threshold on YOLO26x traces an AP@50--EDP Pareto frontier that dominates the evaluated frame-skipping baselines (FixedSkip-$N$, CTD, Statues) within the target operating region ($\mathrm{AP@50}\geq0.50$).

\smallskip
\textbf{Contributions.} This work presents four contributions:

\begin{itemize}
    \setlength{\itemsep}{0.5em}
    \item \textbf{System design.} \Albireo{} is, to our knowledge, the first detector-agnostic, codec-free, content-aware frame-skipping system for edge video object detection that requires no detector modification or detector retraining, is not tied to fixed-camera assumptions, and is designed to improve the accuracy--energy--latency tradeoff of off-the-shelf detectors on commercial edge SoCs. (\Cref{sec:albireo})

    \item \textbf{Forward-looking uncertainty trigger.} We show that the Kalman covariance trace, computed from the current object-state covariance, is a more effective skip signal than the Mahalanobis-against-last-detection signal used by prior work. Under our BDD100K MOT evaluation, this trigger preserves higher AP@50 than CTD at similar skip rates. (\Cref{sec:kf,sec:eval_baselines})

    \item \textbf{Detector-flicker rescue.} We identify detector flicker as a frequent source of missed object outputs on inference frames and introduce a Kalman-rescue rule that preserves confirmed object states through brief detector misses. Our ablation study shows that rescue is the largest AP@50 contributor among \Albireo{}'s components, raising AP@50 from the no-rescue variants' $0.572$--$0.578$ range to $0.617$. (\Cref{sec:rescue,sec:eval_framestate,sec:eval_ablation})

    \item \textbf{Empirical accuracy--energy--latency gains.} Across three detectors and two Jetson platforms, \Albireo{} keeps AP@50 within $1.2$\,pp of Vanilla per-frame inference while reducing energy by $12.1$--$17.6\%$. On the primary YOLO26x configuration, the default operating point improves over Vanilla on all three axes---AP@50, energy, and latency---and the uncertainty-threshold sweep forms the upper AP@50--EDP Pareto frontier over the evaluated frame-skipping baselines (FixedSkip-$N$, CTD, Statues) within the target operating region $\mathrm{AP@50} \geq 0.50$. (\Cref{sec:eval_pareto,sec:eval_baselines,sec:eval_agnostic})
\end{itemize}

The rest of the paper is organized as follows. \Cref{sec:background} motivates our design by discussing shortcomings of existing approaches. \Cref{sec:albireo} describes the system. \Cref{sec:evaluation} presents our experimental evaluation. \Cref{sec:discussion} discusses deployment implications, composability, and limitations. \Cref{sec:related_work} surveys the broader literature, and \Cref{sec:conclusion} concludes.

\section{Background and Motivation}
\label{sec:background}

This section provides the background and motivation for \Albireo{}. 
We first describe the detector families, dataset, and edge-SoC energy model used in our setting (\Cref{sec:edge_stack}), then summarize the Kalman filter primitive reused by our skip oracle (\Cref{sec:kf_bg}). We then show why naive frame skipping is unreliable for video object detection (\Cref{sec:naive_skip}) and explain why existing adaptive alternatives are insufficient for drop-in edge deployment (\Cref{sec:gap}).

\subsection{Edge Inference: Detectors, Camera Motion, and Energy}
\label{sec:edge_stack}

\textbf{Detectors.} The detectors targeted in this work span two common architectural families. \emph{Single-stage CNN detectors}, such as the YOLO family, are widely used in edge deployment because of their favorable latency--accuracy tradeoff, with recent variants further simplifying inference through Distribution Focal Loss (DFL) removal and end-to-end Non-Maximum Suppression (NMS)-free prediction~\cite{khanam2024yolo11,sapkota2025yolo26}. 
\emph{Transformer-based detectors} (DETR~\cite{carion2020detr} and its successors) replace anchors and NMS with set prediction trained via Hungarian matching; RF-DETR~\cite{robinson2026rfdetr} improves this line using neural architecture search for real-time accuracy--latency tradeoffs. 
Both families expose the interface needed by a wrapper system: class-labeled bounding boxes with confidence scores. 
This makes it possible to control detector invocation without modifying the detector architecture or retraining it.

\smallskip
\textbf{Camera motion.} Edge video object detection appears in both fixed-camera settings~\cite{contoli2024dipm,nishimura2022sdof}, such as surveillance and traffic-monitoring nodes, and moving-camera settings~\cite{tran2023fast,yu2020bdd100k}, such as dashcams, autonomous-vehicle cameras, and mobile robots. 
Fixed-camera video often contains strong pixel-level redundancy because the static background regions remain spatially aligned across consecutive frames. 
Moving-camera, or \emph{ego-motion}, video is a harder stress test for frame skipping. As the platform moves, the entire scene shifts continuously, so pixel-level frame comparisons can report large changes, even when the object-level scene is stable.  Methods that rely on frame similarity or background subtraction, therefore, become unreliable under ego-motion (\Cref{sec:gap}). 
\Albireo{} is not tied to a fixed-camera assumption because it reasons over object-level spatial and temporal behavior rather than raw pixel differences. In this paper, we evaluate \Albireo{} on ego-motion video to test this harder deployment regime.

\smallskip
\textbf{Edge SoC energy model.} Edge video analytics typically runs on heterogeneous SoCs that combine GPU acceleration with CPU-side control logic under tight power and thermal budgets. In object-detection pipelines, the detector forward pass is the dominant GPU workload, while bookkeeping, association, and scheduling logic usually execute on the CPU. Prior component-level studies of Jetson-class platforms show that DNN inference energy is strongly shaped by GPU activity and frequency~\cite{dutt2023evaluating,alrafi2024uncovering,holly2020profiling}. This makes detector invocation the natural control point for energy optimization. Skipping a detector frame gates expensive GPU work, while the remaining CPU bookkeeping must be cheap enough not to erase the savings. \Albireo{} therefore targets frame-level detector gating rather than modifying the detector architecture itself.

\subsection{Kalman Filters for Per-Object Temporal State}
\label{sec:kf_bg}

A Kalman filter is a linear-Gaussian state estimator that maintains a state vector $\mathbf{x}$ and a covariance matrix $P$. 
At each time step, \emph{predict} propagates $(\mathbf{x}, P)$ through a linear dynamics model and adds process noise $Q$, increasing uncertainty; \emph{update} fuses an incoming measurement, with measurement noise $R$, through the Kalman gain, reducing uncertainty. 
After repeated predict steps without an update, the covariance $P$ grows, providing a quantitative measure of how stale the current estimate has become.

Kalman filters are commonly used inside tracking-by-detection systems such as SORT~\cite{bewley2016sort} and DeepSORT~\cite{wojke2017deepsort}. 
In that setting, one filter is maintained per object, and detections are associated with the existing object states using an assignment step, often via the Hungarian algorithm~\cite{kuhn1955hungarian} with IoU or embedding-based costs. The usual goal of this approach is to identify preservation: object states coast through brief detector misses and reattach when a matching detection appears. 
In standard SORT-style pipelines, covariance is used mainly for association gating rather than as a system-level control signal. 
\Albireo{} uses the same lightweight per-object state internally, but for a different purpose: it treats covariance growth as a forward-looking signal for deciding when detector inference is needed. 
The output remains detector-style bounding boxes; the Kalman state is a scheduling mechanism, not the evaluation target. Rescue further reuses the same object state to emit predicted boxes that preserve detection recall through brief detector misses, and the empty-scene screen covers object-free scenes where no
covariance signal exists; together these mechanisms form a detector-agnostic, codec-independent, content-aware frame-skipping system
(\Cref{sec:albireo}).
\Cref{sec:kf} develops the trigger formally.

\subsection{Naive Frame Skipping Fails}
\label{sec:naive_skip}

Video streams exhibit substantial temporal redundancy. Consecutive frames often share most of their visual content.  Prior work has exploited temporal structure for efficient video processing~\cite{bobick2001temporal,chen2018scaletime,habibian2021skipconv,liu2018mobile}.  This makes frame skipping attractive. Run the detector every $N$th frame and reuse the previous output on intermediate frames. 
However, fixed-cadence skipping is content-blind. It cannot know when a new object enters the frame, an occlusion clears, or an object accelerates, so it often skips exactly when a fresh detector output is needed. 
In our evaluation using BDD100K MOT and YOLO26x, even FixedSkip-2 loses $8.6$\, pp AP@50 despite skipping only every other frame.  More aggressive fixed skips degrade accuracy further (\Cref{sec:eval_baselines}). 
The design problem is therefore not whether video redundancy exists, but how to decide \emph{which} frames can be skipped safely. That decision must depend on scene content and object-level spatial and temporal state, not on a fixed cadence.

\subsection{Existing Adaptive Approaches Are Insufficient}
\label{sec:gap}

Prior work on efficient video object detection can be grouped into four broad groups.
Across these groups, existing methods each miss at least one requirement for drop-in edge deployment: detector-agnostic operation, no detector retraining, codec-free execution, robustness to camera motion, and energy efficiency on commercial edge SoCs.

\smallskip
\textbf{Temporal feature propagation} methods propagate intermediate feature maps from a detector's neural-network backbone across sparse keyframes using optical flow or learned aggregation~\cite{zhu2017dff,zhu2017fgfa,zhu2018high,shi2024yolov}. 
They reduce per-frame computation, but the propagation module is coupled to the detector backbone and typically requires architecture-specific retraining.

\smallskip
\textbf{Pixel-level frame differencing} skips detection when consecutive frames appear visually similar~\cite{kim2024statues}. 
This can work for fixed-camera surveillance, where the background remains aligned, but it is unreliable under ego-motion because the whole image changes as the camera moves.

\smallskip
\textbf{Learned and confidence-based gating} methods decide when to invoke detection using either a trained scheduler or a runtime confidence signal~\cite{ding2024ctd,luo2019detect,arefeen2022framehopper,xu2022litereconfig}. 
Learned policies require offline training or tuning tied to a detector, dataset, target domain, or platform. 
CTD~\cite{ding2024ctd} is the closest alternative that does not retrain the detector: it uses a Mahalanobis-distance confidence score relative to the most recent detection, but this reference becomes stale across consecutive skipped frames, and the work does not evaluate energy efficiency on commercial edge SoCs.

\begin{table}[t]
\centering
\caption{Deployment properties of frame-skipping and adaptive inference methods for video object detection.}
\label{tab:rw_comparison}
\footnotesize
\setlength{\tabcolsep}{3.2pt}
\begin{tabular}{lcccccc}
\hline
\textbf{Method} & \textbf{DA} & \textbf{NDR} & \textbf{EM} & \textbf{CF} & \textbf{EE} & \textbf{POA} \\
\hline
DFF~\cite{zhu2017dff}                  & \texttimes & \texttimes & \checkmark & \checkmark & \texttimes & \texttimes \\
SDOF-Tracker~\cite{nishimura2022sdof}  & \texttimes & \checkmark & \texttimes & \checkmark & \texttimes & \texttimes \\
Tran et al.~\cite{tran2023fast}        & \checkmark & \texttimes & \checkmark & \texttimes & \texttimes & \texttimes \\
Skip-Conv~\cite{habibian2021skipconv}  & \texttimes & \texttimes & \checkmark & \checkmark & \texttimes & \texttimes \\
PASS~\cite{zhou2024pass}               & \texttimes & \texttimes & \checkmark & \checkmark & \checkmark & \texttimes \\
Statues~\cite{kim2024statues}          & \checkmark & \checkmark & \texttimes & \checkmark & \checkmark & \texttimes \\
CTD~\cite{ding2024ctd}                 & \checkmark & \checkmark & \checkmark & \checkmark & \texttimes & \checkmark \\
Detect or Track~\cite{luo2019detect}   & \texttimes & \texttimes & \checkmark & \checkmark & \texttimes & \texttimes \\
FrameHopper~\cite{arefeen2022framehopper} & \checkmark & \texttimes & \checkmark & \checkmark & \texttimes & \texttimes \\
Henning et al.~\cite{henning2023framedrop} & \checkmark & \checkmark & \checkmark & \checkmark & \checkmark & \texttimes \\
Contoli et al.~\cite{contoli2024dipm}  & \checkmark & \checkmark & \texttimes & \checkmark & \checkmark & \texttimes \\
Compressed-domain~\cite{wang2019compressed,wang2025mayolo} & \checkmark & \texttimes & \checkmark & \texttimes & \texttimes & \texttimes \\
\hline
\textbf{\Albireo{} (ours)}             & \checkmark & \checkmark & \checkmark & \checkmark & \checkmark & \checkmark \\
\hline
\end{tabular}

\vspace{1mm}
\raggedright
\scriptsize
\emph{Columns:} DA = detector-agnostic; NDR = no detector retraining; EM = ego-motion capable; CF = codec-free; EE = energy efficiency evaluated on commercial edge SoCs; POA = per-object adaptive.
\end{table}

\smallskip
\textbf{Compressed-domain methods} reuse codec motion vectors and residuals to propagate detections between keyframes~\cite{wang2019compressed,wang2025mayolo,duche2026compressed,yao2026comprivdet}. 
These methods are useful when compressed bitstreams are available, but they depend on codec metadata that is not exposed in raw camera pipelines before encoding.
The codec keyframe schedules are also chosen to improve
compression, not to determine when the object detector should run.

\Cref{tab:rw_comparison} summarizes this gap across the closest frame-skipping and adaptive-inference methods. 
\Albireo{} targets the gap with a software wrapper that uses per-object uncertainty to gate off-the-shelf detectors without fixed-camera or codec assumptions. 
The next section describes its uncertainty trigger, rescue mechanism, and empty-scene screen.

\section{\Albireo{} System Design}
\label{sec:albireo}

\Albireo{} wraps an off-the-shelf object detector with three mechanisms, each targeting a failure mode of naive frame skipping (\Cref{sec:background}): i) a \emph{forward-looking uncertainty trigger} that gates detector invocation using per-object Kalman covariance growth, ii) a \emph{detector-flicker rescue} mechanism that keeps confirmed object states alive when the detector briefly misses them, and iii) a lightweight \emph{empty-scene screen} that avoids full detector calls on objectless frames. 
These mechanisms require no detector modifications or detector retraining. The uncertainty threshold $u$ serves as the primary knob for selecting the accuracy--efficiency operating point. 
We describe the three mechanisms in turn, then unify them in the per-frame algorithm of \Cref{sec:framestates}.

\Albireo{} targets causal, online, single-stream edge inference: every
decision uses only the current frame and state carried from earlier frames,
and future frames are never accessed. We make no hard real-time or
safety-critical guarantees. In our evaluation, recorded clips are replayed
to enable reproducible accuracy, latency, and power measurement
(\Cref{sec:evaluation}).

\subsection{Per-Object Kalman Filter and Uncertainty Trigger}
\label{sec:kf}

Each active object state is represented by a 10-dimensional Kalman filter in normalized image coordinates,
\[
\mathbf{x} =
[\,c_x,\, c_y,\, w,\, h,\, \dot c_x,\, \dot c_y,\, \dot w,\, \dot h,\, \ddot c_x,\, \ddot c_y\,]^\top ,
\]
where $(c_x,c_y)$ is the box center and $(w,h)$ is the box size. 
The model uses constant-acceleration dynamics for the box center, allowing the state to represent speed changes such as acceleration, deceleration, or abrupt motion. Additionally, it uses constant-velocity dynamics for the box size, allowing width and height to change smoothly as objects move toward or away from the camera.  
Coordinates are normalized by frame dimensions, making the state representation independent of input resolution. 
We use the standard linear-Gaussian predict and update equations as in SORT~\cite{bewley2016sort}; the transition matrix $F \in \mathbb{R}^{10 \times 10}$, observation matrix $H \in \mathbb{R}^{4 \times 10}$, and noise covariances $Q,R$ are fixed implementation parameters.

\Albireo{} uses the projected covariance trace as its uncertainty signal:
\begin{equation}
\tau = \mathrm{tr}\!\left(H P H^\top\right),
\label{eq:trace}
\end{equation}
where $H$ selects the observed box coordinates $(c_x,c_y,w,h)$ from the state. 
Predict steps increase $\tau$ because no new detector measurement has constrained the object state (when the detector is skipped, $P \leftarrow F P F^\top + Q$, and uncertainty grows); update steps reduce $\tau$ after a detector measurement is incorporated. 
After the per-frame prediction step, \Albireo{} skips detector invocation when all active object states satisfy $\tau < u$, subject to the safety caps below (\Cref{alg:step}). 
The threshold $u$ is the primary control knob: lower values invoke the detector more often, while higher values allow longer prediction-only stretches.

This signal is forward-looking because it is computed from the filter state and covariance, not by comparing the current prediction to an old detector output. 
This differs from CTD~\cite{ding2024ctd}, which uses a Mahalanobis-distance score relative to the most recent detection; as consecutive skips accumulate, that reference detection becomes increasingly stale. 
In \Albireo{}, uncertainty increases during each Kalman prediction step when no detector measurement is available, so the trigger reflects how long the object state has evolved without being corrected by a new detection.

With a fixed $Q$, object velocity affects the predicted box
location---the state mean---but does not impact the growth rate of the covariance;
stationary and moving objects accrue the same model-based uncertainty
between corrections. $\tau$ reflects elapsed time without
correction rather than motion magnitude: for slow objects, this may invoke
the detector more often than necessary, impacting energy, while for fast
objects, the covariance may not grow fast enough. The caps below
independently bound consecutive prediction-only frames; motion-adaptive
process noise is future work.

Two safety caps bound worst-case behavior. 
Let $\Delta T_t$ be the number of consecutive frames since object state $t$ was last updated by a real detector measurement. 
We force detector invocation if any active state reaches $T_{\max}=15$ prediction-only frames, and we also invoke the detector at least once every 30 frames while active states exist. 
The first cap bounds drift during long prediction stretches; the second bounds detector-call latency, even when the covariance remains below the threshold. 
When no active object states exist, discovery of newly appearing objects is handled by the empty-scene screen described in \Cref{sec:erd}.

\subsection{Detector-Flicker Rescue}
\label{sec:rescue}

A single-image detector scores each frame independently. 
An object confidently detected at frame $t$ may fall below the confidence threshold at frame $t{+}1$ because of partial occlusion, lighting change, motion blur, or scale change, even though the object remains visible. 
We refer to this transient miss as \emph{detector flicker}. 
Without a rescue rule, a confirmed object state that misses one detector frame can be killed, then later re-created as a tentative state when the detector recovers. 
If tentative states require multiple hits before they are emitted, the object can disappear from the output for several frames around the flicker, reducing detection quality despite invoking the detector.

\Albireo{} intercepts this failure case on inference frames. 
After matching detector outputs to existing object states, we examine confirmed states that remain unmatched. 
An unmatched confirmed state is rescued if two conditions hold: (i) its most recent real detector match had confidence $\geq c_{\mathrm{rescue}} = 0.5$ (matching the detector confidence threshold used in our evaluation), and (ii) it has been rescued fewer than $R_{\max}=2$ consecutive times.  These conservative defaults are intended to bridge short detector flicker without allowing long unsupported prediction: $c_{\mathrm{rescue}}$ restricts rescue to recently high-confidence objects, and $R_{\max}$ bounds drift by limiting rescue to at most two consecutive frames.
A rescued state emits its Kalman-predicted bounding box for the current frame, but it does not receive a Kalman update because no real detector measurement was observed (\Cref{alg:step}). 
Thus, its uncertainty $\tau$ continues to grow across rescued frames, forcing a future detector invocation if the state remains unsupported. 
A real detector match resets the rescue counter; after $R_{\max}$ consecutive unmatched inference frames, the state is removed normally. For deployments with different motion statistics, $R_{\max}$ should be interpreted as a short temporal budget rather than a universal frame count. In our pilot sweeps, longer rescue windows such as 5 or 8 frames reduced accuracy by allowing stale boxes to persist.

\subsection{Empty-Scene Screening}
\label{sec:erd}

When no active object states exist, the uncertainty trigger has no per-object covariance to consult. 
This occurs at the start of a clip, after objects leave the scene, and during long objectless stretches. A conservative fallback would run the full detector on every frame until a new object appears, paying full inference cost even when the scene remains empty. 
\Albireo{} instead uses a lightweight binary scene classifier, the empty-road detector (ERD)~\cite{liu2022erd}, to decide whether full detection is needed. 
ERD is a five-block CNN with $\sim$14k parameters and $\sim$1.3\,ms GPU latency that labels a frame as empty or non-empty. We fine-tune this classifier on BDD100K training data to match the ego-motion domain. This one-time adaptation is lightweight because ERD has only $\sim$14k parameters; the object detector itself is not modified or retrained.

ERD is invoked only when the active object-state set is empty. 
If ERD predicts \emph{empty}, \Albireo{} skips the frame without running the detector. 
If ERD predicts \emph{non-empty}, \Albireo{} invokes the detector and initializes new object states from the detector outputs (\Cref{alg:step}). 
Because ERD is not evaluated on frames that already contain active object states, its overhead is proportional to the no-active-state fraction rather than to the full video length.

\begin{algorithm}[t]
\small
\caption{\Albireo{} per-frame step.}
\label{alg:step}
\begin{algorithmic}[1]
\REQUIRE frame; object-state set $\mathcal{T}$
\ENSURE emitted boxes; frame state $s \in \{E,P,E{+}I,I,AI\}$

\FORALL{states $t \in \mathcal{T}$}
  \STATE Predict $(\mathbf{x}_t,P_t)$ using $F,Q$; remove $t$ if its predicted center leaves the frame.
\ENDFOR

\IF{$\mathcal{T}$ has no active states}
  \IF{$\mathrm{ERD}(\mathrm{frame})=\mathrm{empty}$}
    \RETURN $\emptyset$, $E$
  \ELSE
    \STATE $D \leftarrow \mathrm{Detector}(\mathrm{frame})$; initialize tentative states from $D$
    \RETURN $D$, $E{+}I$
  \ENDIF
\ENDIF

\STATE $\tau_{\max} \leftarrow \max_{t\in\mathcal{T}}\mathrm{tr}(HP_tH^\top)$;\quad
       $\Delta T_{\max} \leftarrow \max_{t\in\mathcal{T}}\Delta T_t$

\IF{$\tau_{\max}<u$ \textbf{and} $\Delta T_{\max}<T_{\max}$ \textbf{and} not periodic-cap frame}
  \RETURN predicted boxes from $\mathcal{T}$, $P$
\ENDIF

\STATE $D \leftarrow \mathrm{Detector}(\mathrm{frame})$
\STATE $M \leftarrow \mathrm{TwoStageMatch}(D,\mathcal{T})$ \COMMENT{IoU, then Mahalanobis}
\STATE Update matched states using $M$; promote tentative states as appropriate.
\STATE $r \leftarrow 0$

\FORALL{unmatched confirmed state $t$}
  \IF{$\mathrm{last\_conf}_t \geq c_{\mathrm{rescue}}$ \textbf{and} $\rho_t < R_{\max}$}
    \STATE Emit predicted box for $t$; $\rho_t \leftarrow \rho_t+1$; $r \leftarrow r+1$
  \ELSE
    \STATE Remove $t$
  \ENDIF
\ENDFOR

\IF{$r>0$}
  \RETURN merged detector and rescued boxes, $AI$
\ELSE
  \RETURN detector boxes, $I$
\ENDIF
\end{algorithmic}
\end{algorithm}
In a non-empty scene, a newly appearing object has no Kalman
state, so its discovery may be delayed until the next detector invocation;
the periodic caps (\Cref{sec:kf}) bound this delay, and an existing
object's growing uncertainty may trigger detection earlier. A reappearing
object is handled through its existing state while it remains active.
Delayed discoveries on skipped frames are counted as false negatives in
AP@50, so their cost is reflected in our reported accuracy.

\subsection{Frame-State Taxonomy and Algorithm}
\label{sec:framestates}

\begin{figure}[t]
\centering
\begin{tikzpicture}[
  font=\footnotesize,
  >={Latex[length=1.6mm]},
  state/.style={circle, draw, very thick, minimum size=10mm,
                inner sep=0pt, align=center, font=\footnotesize\bfseries},
  state-E/.style={state, fill=teal!25},
  state-P/.style={state, fill=magenta!20},
  state-EI/.style={state, fill=orange!25},
  state-I/.style={state, fill=violet!20},
  state-AI/.style={state, fill=red!25},
  edge-lbl/.style={font=\scriptsize, fill=white, inner sep=1pt}
]
  \node[state-E]  (E)  at (0, 3.2)   {E};
  \node[state-EI] (EI) at (3.8, 3.2) {E+I};
  \node[state-I]  (I)  at (1.9, 1.2) {I};
  \node[state-AI] (AI) at (4.8, 1.2) {AI};
  \node[state-P]  (P)  at (1.9,-0.8) {P};

  \draw[->] (4.9, 3.6) -- (EI);

  \draw[->] (E)  to[bend left=18]
    node[edge-lbl, above]{scene non-empty} (EI);
  \draw[->] (EI) to[bend left=18]
    node[edge-lbl, below]{new tracks die} (E);

  \draw[->] (EI) to
    node[edge-lbl, sloped, above=1pt]{tracks confirmed} (I);

  \draw[->] (I)  to[bend left=18]
    node[edge-lbl, above]{rescue} (AI);
  \draw[->] (AI) to[bend left=18]
    node[edge-lbl, below]{clean match} (I);

  \draw[->] (I) to[bend right=20]
    node[edge-lbl, left=1pt]{trigger silent} (P);
  \draw[->] (P) to[bend right=20]
    node[edge-lbl, right=1pt]{trigger fires} (I);

  \draw[->] (I) to[bend left=30]
    node[edge-lbl, sloped, below, pos=0.45]{tracks lost} (E);

  \draw[<->] (P) to[bend right=30]
    node[edge-lbl, sloped, below=1pt, pos=0.45]{trigger fires + rescue} (AI);
\end{tikzpicture}
\vspace{0.1cm}
\hrule
\vspace{0cm}
\caption{\textbf{Frame-state transitions across consecutive frames.} Edges show the events that move execution between ERD skipping, prediction-only, normal inference, and rescue-augmented inference states. AI follows the same next-frame routing as I; self-loops are omitted.}
\label{fig:state_machine}
\end{figure}
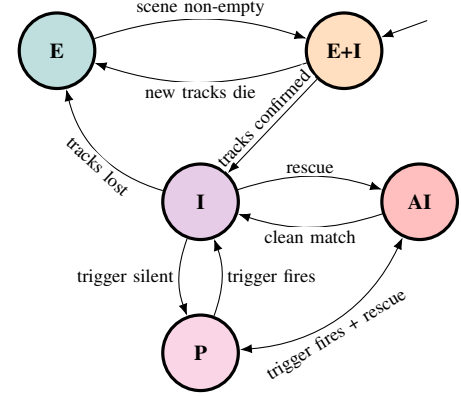

\Albireo{} combines the three mechanisms above and assigns each processed frame to one of the five mutually exclusive states, summarized in \Cref{tab:framestates}. 
Each state records which computation produced the frame output: ERD skipping, prediction-only output, ERD-triggered inference, normal detector inference, or detector inference augmented by rescue.
\Cref{fig:state_machine} illustrates how execution moves between these states across consecutive frames as scenes become empty or non-empty, uncertainty triggers fire, detections match, or rescue is needed. 
\Cref{alg:step} gives the exact per-frame routing logic, combining Kalman prediction, boundary pruning, empty-scene screening, uncertainty-triggered skipping, detector matching, and rescue in a single pass. 
We use the same state taxonomy in the evaluation to attribute energy savings and accuracy changes to specific runtime behaviors.

In terms of per frame overhead, excluding the detector call itself, \Albireo{}'s overhead is modest: prediction costs $O(n)$ over $n$ active object states, forming the association cost matrix costs $O(nm)$ for $m$ detections, and Hungarian matching is $O(\max\{n,m\}^{3})$ in the worst case; the empty-scene screen runs only when no active object states exist.

\begin{table}[t]
\centering
\caption{Frame states used by \Albireo{}. Each frame is assigned exactly one state.}
\label{tab:framestates}
\footnotesize
\setlength{\tabcolsep}{4pt}
\begin{tabular}{cll}
\hline
\textbf{State} & \textbf{Name} & \textbf{Frame output / computation} \\
\hline
E    & ERD-Empty            & ERD only; detector skipped \\
P    & KF-Predict           & Predicted boxes; detector skipped \\
E+I  & ERD-triggered Infer. & ERD then detector \\
I    & Inference            & Detector, match, KF update \\
AI   & Augmented Inference  & Detector, match/update, rescue output \\
\hline
\end{tabular}
\end{table}

\section{Evaluation}
\label{sec:evaluation}

\subsection{Experimental Setup}
\label{sec:eval_setup}

\textbf{Dataset.} We evaluate \Albireo{} on the multi-object tracking (MOT) validation split of the BDD100K dataset~\cite{yu2020bdd100k}, a benchmark of 200 dashcam clips collected on real roads. Each clip is approximately 40\,s long at the BDD100K MOT label rate of 5\,fps, yielding $\sim$40{,}000 annotated frames per configuration. We restrict evaluation to the seven object classes that have per-frame bounding-box annotations in the MOT split (\texttt{car}, \texttt{truck}, \texttt{bus}, \texttt{pedestrian}, \texttt{rider}, \texttt{bicycle}, \texttt{motorcycle}); all other classes are excluded. We use BDD100K MOT as a representative stress test for ego-motion driving workloads: it contains real dashcam video with moving cameras, changing traffic density, object entry/exit, occlusion, and scale changes, which are the conditions that stress frame-skipping policies.

\smallskip
\textbf{Detectors.} To support the detector-agnostic claim, we evaluate \Albireo{} as a wrapper around three architecturally distinct off-the-shelf detectors: YOLO11x~\cite{khanam2024yolo11} and YOLO26x~\cite{sapkota2025yolo26} (single-stage CNN detectors) and RF-DETR-Large~\cite{robinson2026rfdetr} (a transformer-based detector). All three are used at the same confidence threshold ($c=0.50$); class outputs are filtered to the seven BDD100K MOT classes. We treat YOLO26x as the primary detector for headline tables and figures, and report YOLO11x and RF-DETR-Large to demonstrate that the pattern of accuracy and energy savings is consistent across detector families.

\smallskip
\textbf{Platforms.} We deploy on two consecutive generations of NVIDIA Jetson edge SoC: AGX Orin (Ampere GPU, 12-core ARM Cortex-A78AE CPU 32\,GB LPDDR5, 204.8\,GB/s, 60\,W TDP)~\cite{nvidia_orin_manual} and AGX Thor (Blackwell GPU, 14-core ARM Neoverse CPU, 128\,GB LPDDR5x, 273\,GB/s, 130\,W TDP)~\cite{nvidia_thor_trm_2025}. The two boards span a $\sim$1.3$\times$ memory-bandwidth gap and a $\sim$2.2$\times$ TDP gap, allowing us to test whether \Albireo{}'s savings transfer across SoC generations and power envelopes. Orin runs JetPack 6.2 (CUDA 12.6, PyTorch 2.11) and Thor runs JetPack 7.1 (CUDA 13.2, PyTorch 2.11). YOLO11x and YOLO26x are loaded via Ultralytics 8.4.39, and RF-DETR-Large via rfdetr 1.6.4; all three run directly under PyTorch without TensorRT conversion or other platform-specific compilation.

\smallskip
\textbf{Baselines.} We compare \Albireo{} against five families of frame-skipping baselines. \emph{Vanilla} runs the detector on every frame. \emph{FixedSkip-$N$} ($N\!\in\!\{2,3,5\}$) runs the detector every $N$th frame and reuses the previous detection on intermediate frames (this is the simplest, content-blind frame-skipping strategy). \emph{CTD}~\cite{ding2024ctd} is a Python reimplementation of the Mahalanobis-distance trigger of Ding et al., swept across confidence thresholds $\{0.30, 0.50, 0.70, 0.90\}$. \emph{Statues}~\cite{kim2024statues} is a Python reimplementation of the pixel-level frame-differencing scheme of Kim et al., swept across connected-component-size thresholds $\{25, 75, 200, 500\}$, at a fixed pixel-difference threshold of 30. \emph{ERD-only} runs the empty-road classifier of Liu and Kang~\cite{liu2022erd} on every frame and skips detection when the scene is classified empty, with no Kalman tracking. For \Albireo{} we report the default operating point ($u=3\!\times\!10^{-4}$) and a sweep over the uncertainty trigger $u \in \{3\!\times\!10^{-4}, 5\!\times\!10^{-4}, 7\!\times\!10^{-4}, 1\!\times\!10^{-3}\}$. Both \Albireo{} and ERD-only use the same ERD model, initialized from Liu and Kang~\cite{liu2022erd} and fine-tuned on BDD100K training clips to classify frames as empty or non-empty. We do not include SORT~\cite{bewley2016sort} or DeepSORT~\cite{wojke2017deepsort} as baselines: they are trackers rather than adaptive detector-invocation methods---the detector still runs on every frame---so including them would measure per-frame detection plus tracking overhead rather than isolating the
detector-scheduling decision. CTD is the direct Kalman-based scheduling
baseline. The BDD100K MOT validation clips used for evaluation are not used for ERD fine-tuning.

\smallskip
\textbf{Metrics.} We report seven metrics. Detection accuracy is measured by AP@50 (VOC-style, IoU 0.5) and mAP@50:95 (COCO-style, IoU swept 0.5--0.95). Efficiency is measured by average frame time (ms), total clip energy (J), and per-frame energy-delay product (EDP). 
For a clip with energy $E_{\mathrm{clip}}$, duration $T_{\mathrm{clip}}$, and $N$ frames, per-frame EDP is $(E_{\mathrm{clip}}T_{\mathrm{clip}})/N$, reported in J$\cdot$ms/frame. 
Runtime behavior is measured by skip rate (\%) and the five-state frame breakdown defined in \Cref{sec:framestates}. 
For AP computation, \Albireo{} assigns Kalman-predicted and rescue boxes the most recent real detector confidence stored in the object state---detector-produced boxes retain their original detector scores. FixedSkip baselines reuse both the previous box and its previous confidence on skipped frames.
We treat AP@50 as the primary accuracy metric because prediction-only frames can preserve object-level matches while producing boxes that are less precise at higher IoU thresholds (e.g., $\geq 0.75$), even when the underlying object identity is correct.

\smallskip
\textbf{Power and energy measurement.} Power is sampled every 100\,ms using a per-clip \texttt{tegrastats} subprocess. 
We parse platform-specific rails into GPU, CPU, and IO/MEM domains. 
Clip energy is computed by trapezoidal integration over raw power samples. This preserves the bursty power profile created by detector-inference frames and skipped frames.
Total energy uses the top-level board input rail; per-domain rails are used only for the GPU/CPU/IO breakdown in \Cref{sec:eval_perdomain}. 
On Thor, grouped per-domain rails sum to approximately 7\,W less than the top-level input, reflecting unmapped rails.

\smallskip
\textbf{Protocol.} For each detector--platform pair, all systems process the same 200 clips with seed 42 and identical clip orderings. 
Vanilla and \Albireo{} share a single per-clip Python orchestrator for paired timing and power measurement, so both systems are measured under closely matched conditions. 
Skip decisions are deterministic given the detector outputs, seed, and clip order; we verify this by checking that \Albireo{}'s frame-state percentages match across Orin and Thor to within rounding (\Cref{sec:eval_framestate}).


\subsection{Pareto Dominance Across Three Axes}
\label{sec:eval_pareto}

On the primary YOLO26x configuration, \Albireo{}'s default operating point ($u=3\!\times\!10^{-4}$) Pareto-dominates Vanilla per-frame inference across AP@50, energy, and latency simultaneously. \Cref{tab:main} reports the headline numbers for YOLO26x on Thor and Orin: \Albireo{} gains $+0.8$\,pp~AP@50 (a $1.2\%$ relative improvement) while reducing energy per clip by $17.6\%$ on Thor ($14.4\%$ on Orin) and average frame time by $8.8\%$ on Thor ($13.7\%$ on Orin). The per-frame EDP drops by $24.9\%$ on Thor and by $26.1\%$ on Orin. These improvements occur together: each set of parameters (AP@50, energy, frame time) moves in a favorable direction (i.e., higher accuracy, lower energy and latency), so the default \Albireo{} point dominates Vanilla in the strict Pareto sense rather than trading one axis for another.

\begin{table}[t]
\centering
\caption{YOLO26x headline results on BDD100K MOT. Values are means over 200 validation clips.}
\label{tab:main}
\footnotesize
\setlength{\tabcolsep}{4pt}
\begin{tabular}{llccccc}
\hline
\textbf{Plat.} & \textbf{System} & \textbf{AP@50} & \textbf{Energy} & \textbf{ms/fr} & \textbf{EDP} & \textbf{Skip} \\
 &  &  & \textbf{(J)} &  & \textbf{(J$\cdot$ms/fr)} & \textbf{(\%)} \\
\hline
Thor & Vanilla    & 0.609 & 811.5 & 50.9 & 41{,}305 & 0.0 \\
Thor & \Albireo{} & 0.617 & 668.6 & 46.4 & 31{,}023 & 19.3 \\
\multicolumn{2}{r}{$\Delta$} & \textbf{+0.8\,pp} & \textbf{$-$17.6\%} & \textbf{$-$8.8\%} & \textbf{$-$24.9\%} & --- \\
\hline
Orin & Vanilla    & 0.609 & 994.0 & 111.5 & 110{,}831 & 0.0 \\
Orin & \Albireo{} & 0.617 & 851.2 &  96.2 &  81{,}885 & 19.3 \\
\multicolumn{2}{r}{$\Delta$} & \textbf{+0.8\,pp} & \textbf{$-$14.4\%} & \textbf{$-$13.7\%} & \textbf{$-$26.1\%} & --- \\
\hline
\end{tabular}
\end{table}

\begin{figure*}[t]
\centering
\includegraphics[width=\linewidth]{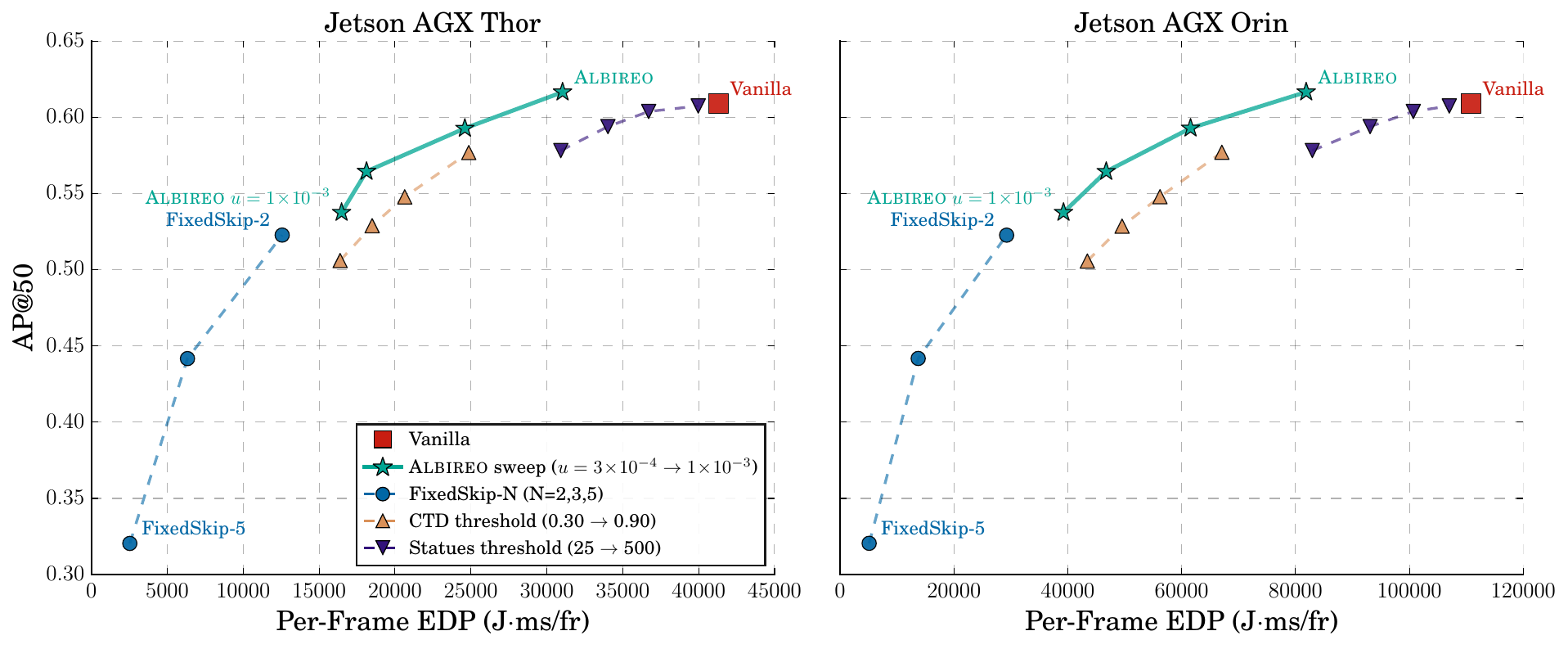}
\vspace{-0.5cm}
\hrule
\vspace{0cm}
\caption{AP@50 versus per-frame energy-delay product (EDP) for YOLO26x on the BDD100K MOT validation split. 
\Albireo{}'s threshold sweep ($u=3\!\times\!10^{-4}$ to $1\!\times\!10^{-3}$) forms the upper Pareto frontier among the evaluated systems in the $\mathrm{AP@50}\geq0.50$ operating region on both Jetson platforms.}
\label{fig:pareto_edp}
\end{figure*}

The uncertainty threshold $u$ is \Albireo{}'s primary control knob. Sweeping $u$ upward (looser trigger, more skipping, lower accuracy) traces a curve in (AP@50, EDP) space. \Cref{fig:pareto_edp} shows this curve alongside every comparable frame-skipping baseline on both platforms. We maintain that the default operating point is not an
accuracy--efficiency trade-off relative to per-frame inference: it improves AP@50, energy and latency simultaneously; the threshold sweep then characterizes the trade-off at more aggressive skip rates.

We focus on $\mathrm{AP@50}\geq0.50$ as the target operating region for skip-policy comparison. 
Accordingly, the main \Albireo{} sweep stops at $u=1\!\times\!10^{-3}$; larger thresholds leave this region by allowing longer prediction-only stretches and larger accuracy loss. 
We also evaluate more aggressive thresholds as stress tests. 
For example, at $u=3\!\times\!10^{-3}$, \Albireo{} reaches $55\%$ energy savings but drops to $\mathrm{AP@50}=0.46$, outside the target operating region.


Within the target operating range $\mathrm{AP@50}\geq0.50$, \Cref{fig:pareto_edp} shows that \Albireo{} forms the upper Pareto frontier among the evaluated systems on both platforms. 
At comparable EDP, baseline methods achieve lower AP@50; at comparable AP@50, they require higher EDP. 
The advantage is largest near the high-accuracy end: on Thor, Vanilla requires $33\%$ more EDP than the default \Albireo{} point, while also achieving lower AP@50. 
Near the lower end of the target range, the margin narrows, but the evaluated FixedSkip and CTD points remain below the \Albireo{} frontier. 
The next subsection breaks down these gaps by baseline family and also reports the corresponding energy and latency comparisons.

\subsection{Comparison to Prior Frame-Skipping Baselines}
\label{sec:eval_baselines}

\textbf{FixedSkip-$N$.} Content-blind skipping loses accuracy quickly as $N$ grows. 
FixedSkip-2 (50\% skip) loses $8.6$\,pp AP@50 (a $14.2\%$ relative drop), FixedSkip-3 loses $16.7$\,pp, and FixedSkip-5 (80\% skip) loses $28.8$\,pp ($47\%$ relative drop). 
The fixed schedule has no awareness of when objects enter the frame, recover from occlusion, or change motion, so it often skips frames where a fresh detector output is needed. 
FixedSkip-3 and FixedSkip-5 do reach lower absolute energy and frame times than \Albireo{}, but only at $\mathrm{AP@50}=0.44$ and $\mathrm{AP@50}=0.32$, respectively, outside the target operating region. 
At the low-AP end of the target range, \Albireo{} at $u=1\!\times\!10^{-3}$ reaches $\mathrm{AP@50}=0.538$, while FixedSkip-2 reaches comparable EDP, but only at the cost of an $\mathrm{AP@50}=0.523$, remaining Pareto-worse by a narrower margin.

\smallskip
\textbf{CTD.} CTD is the closest prior mechanism to \Albireo{} because it also uses a Kalman-based trigger, but \Albireo{} achieves higher AP@50 at comparable skip rates. 
With an approximate skip rate of $40\%$, \Albireo{} ($u=7\!\times\!10^{-4}$) reaches $\mathrm{AP@50}=0.564$, while CTD-0.30 reaches $0.506$, a $5.8$\,pp gap. 
With an approximate skip rate of $31\%$, \Albireo{} ($u=5\!\times\!10^{-4}$) reaches $\mathrm{AP@50}=0.593$, as compared with $0.548$ for CTD-0.70. 
Near $\mathrm{AP@50}\approx0.577$, the closest CTD point (CTD-0.90) requires $\approx16\%$ more EDP than the corresponding point on the \Albireo{} curve, interpolated between $u=5\!\times\!10^{-4}$ and $u=7\!\times\!10^{-4}$. 
The difference is structural: CTD scores confidence relative to the most recent detection, so the reference becomes stale across consecutive skipped frames. CTD uses a Mahalanobis-distance score relative to the most recent detector output; as consecutive skips accumulate, that reference becomes stale. \Albireo{} instead gates inference using covariance growth from the current object state. The trigger reflects how long the state has evolved without measurement. In addition, \Albireo{} addresses two failure modes outside CTD's skip oracle: detector flicker on inference frames, handled by the rescue mechanism, and no-active-object frames, handled by conditional empty-scene screening. These system components, together with measured energy evaluation on commercial Jetson platforms, distinguish \Albireo{} from CTD as a complete edge inference framework rather than only an alternative Kalman confidence score.


\smallskip
\textbf{Statues.} Statues~\cite{kim2024statues} relies on pixel-level frame differencing, which is effective mainly when the camera is fixed. 
Under dashcam ego-motion, frame differencing produces large changes across much of the image, so the connected-component score rarely falls below the skip threshold. 
Across the four thresholds we evaluate, Statues skips only $3.6$--$15.5\%$ of frames. 
Even the most aggressive setting, Statues-500, reaches only $\mathrm{AP@50}=0.578$ at $650$\,J, while \Albireo{} at $u=5\!\times\!10^{-4}$ reaches $\mathrm{AP@50}=0.593$ at $602$\,J. 
At the lowest threshold, Statues-25 gives only marginal improvement over Vanilla because frame-difference preprocessing runs on every frame, with skips triggered on only $3.6\%$ of frames.

\smallskip
\textbf{ERD-only.} ERD-only is a baseline that tests whether scene-level empty-frame screening is sufficient without per-object state. 
Even with the same BDD100K-fine-tuned ERD used by \Albireo{}, running ERD on every frame yields a $7.6\%$ skip rate, though per-clip energy increases by $28\%$ and the average frame time grows by $134\%$ on Thor. 
The reason is straightforward: the ERD CNN runs on all frames, and the full detector still runs on the $\sim$92\% classified as non-empty. 
This result explains why \Albireo{} invokes ERD only when no active object states exist: ERD is useful as a conditional empty-scene screen, not as a per-frame replacement for object-level reasoning.

\subsection{Detector-Agnostic Behavior Across Edge Platforms}
\label{sec:eval_agnostic}

\begin{figure}[t]
\centering
\includegraphics[width=\linewidth]{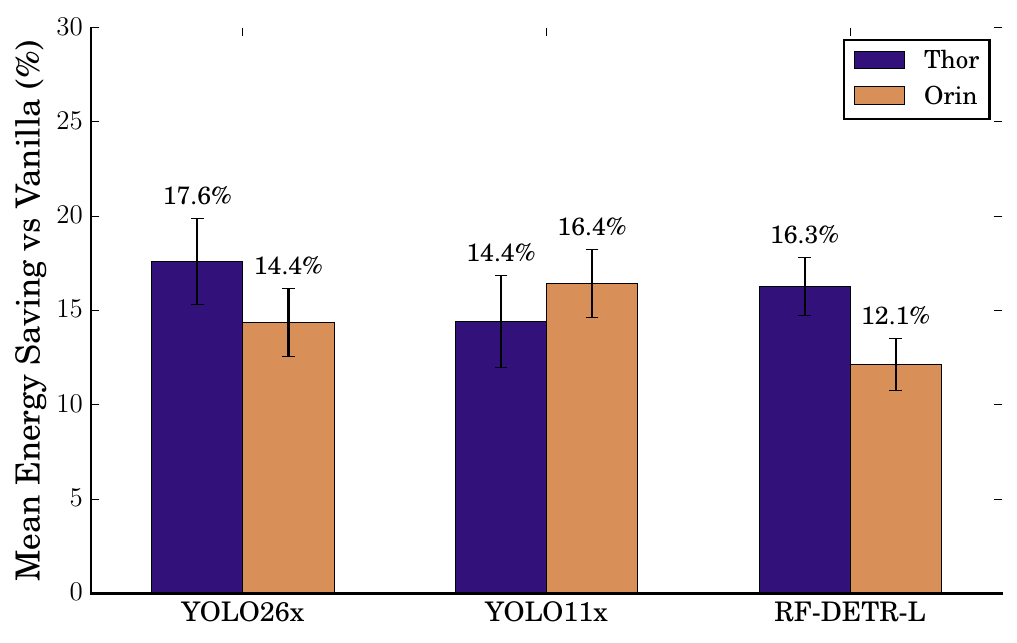}
\vspace{-0.5cm}
\hrule
\vspace{0cm}
\caption{Per-clip energy savings of default \Albireo{} over Vanilla across detector families and Jetson platforms.}
\label{fig:energy_savings}
\end{figure}

\Cref{tab:cross_detector} evaluates \Albireo{}'s detector-agnostic design across three off-the-shelf detectors and two Jetson platforms. 
Across the evaluated detectors, AP@50 remains within $\pm1.2$\,pp of Vanilla while energy savings range from $12.1\%$ on RF-DETR-Large/Orin to $17.6\%$ on YOLO26x/Thor. \Cref{fig:energy_savings} summarizes the energy-savings trend visually. These reductions are statistically significant in all six detector--platform pairs (paired test over the 200 shared clips, $p<0.001$), with \Albireo{} consuming less energy than Vanilla on
$82$--$95\%$ of individual clips. On battery-powered deployments, this
directly extends operating lifetime for the same workload~\cite{taherin2018tsusc, taherin2015stretch}.

\begin{table}[t]
\centering
\caption{Default \Albireo{} relative to Vanilla across detectors and platforms. Values are changes from per-frame inference, averaged over 200 BDD100K MOT validation clips.}
\label{tab:cross_detector}
\footnotesize
\setlength{\tabcolsep}{4pt}
\begin{tabular}{llcccc}
\hline
\textbf{Plat.} & \textbf{Detector} & \textbf{$\Delta$AP@50} & \textbf{$\Delta$mAP} & \textbf{$\Delta$Energy} & \textbf{Skip} \\
 &  & \textbf{(pp)} & \textbf{(pp)} &  & \textbf{(\%)} \\
\hline
Thor & YOLO11x        & $-0.5$ & $-4.7$ & $-14.4\%$ & 21.3 \\
Thor & YOLO26x        & $+0.8$ & $-3.7$ & $-17.6\%$ & 19.3 \\
Thor & RF-DETR-Large  & $-1.1$ & $-4.8$ & $-16.3\%$ & 22.2 \\
\hline
Orin & YOLO11x        & $-0.5$ & $-4.7$ & $-16.4\%$ & 21.3 \\
Orin & YOLO26x        & $+0.8$ & $-3.7$ & $-14.4\%$ & 19.3 \\
Orin & RF-DETR-Large  & $-1.2$ & $-4.8$ & $-12.1\%$ & 22.2 \\
\hline
\end{tabular}
\end{table}

Two trends are worth noting. 
First, for a fixed detector, skip rates are identical across Thor and Orin, to within rounding. 
This is expected because \Albireo{}'s skip decisions depend on detector outputs and object-state uncertainty, not on hardware power or latency measurements. 
Energy savings vary, even on the same skip pattern, because GPU savings and CPU-power responses are platform dependent. 
For a fixed detector, \Albireo{} performs the same per-frame bookkeeping on Thor and Orin, but Orin's coarser CPU DVFS clustering makes this control overhead more visible in CPU power, while Thor's finer-grained CPU complex absorbs easily. 
Thus, the same skipped frames produce different total-energy reductions depending on GPU power, detector latency, and CPU DVFS behavior.
Second, the mAP@50:95 drops cluster around $3.7$--$4.8$\,pp across detectors and platforms. 
This is consistent with the metric behavior discussed in \Cref{sec:eval_setup}: prediction-only frames can preserve AP@50-level matches while producing boxes that are less precise at higher IoU thresholds. 
The similar mAP drop across detectors suggests that the residual high-IoU loss is not primarily a detector-specific failure mode.

\Albireo{}'s benefits also hold while considering the platform power
budget. We further evaluate all three \texttt{nvpmodel} power modes on
each board (YOLO26x, the same 200 clips per mode;
\Cref{tab:power-modes}): total energy savings stay within
$15.3$--$17.7\%$ across a $4.6\times$ power-budget spread, and are
largest at Orin's strict 15\,W budget. The AP@50 gain ($+0.8$\,pp) and
skip rate ($19.3\%$) are identical in every mode, confirming that skip
decisions are content-driven and independent of the power budget.
Investigating the interaction between the power budget and the skip
policy, as well as examining its impact on end-to-end latency, are areas
for future work.

\begin{table}[t]
\centering
\caption{Per-clip energy under each \texttt{nvpmodel}
power mode (YOLO26x, mean over 200 clips per mode). The AP@50 gain
($+0.8$\,pp) and skip rate ($19.3\%$) are identical in every mode.}
\label{tab:power-modes}
\setlength{\tabcolsep}{5pt}
\footnotesize
\begin{tabular}{@{}llccc@{}}
\toprule
\textbf{Board} & \textbf{Mode} & \textbf{Vanilla (J/clip)} &
\textbf{\Albireo{} (J/clip)} & \textbf{$\Delta$Energy} \\
\midrule
Orin & 15\,W  & 1417 & 1166 & $-17.7\%$ \\
Orin & 30\,W  & 1094 &  917 & $-16.2\%$ \\
Orin & 50\,W  &  847 &  714 & $-15.7\%$ \\
\addlinespace
Thor & 70\,W  &  920 &  776 & $-15.7\%$ \\
Thor & 90\,W  &  920 &  779 & $-15.3\%$ \\
Thor & 120\,W &  779 &  654 & $-16.1\%$ \\
\bottomrule
\end{tabular}
\end{table} 

\subsection{Frame-State Behavior}
\label{sec:eval_framestate}

\begin{figure}[t]
\centering
\includegraphics[width=\linewidth]{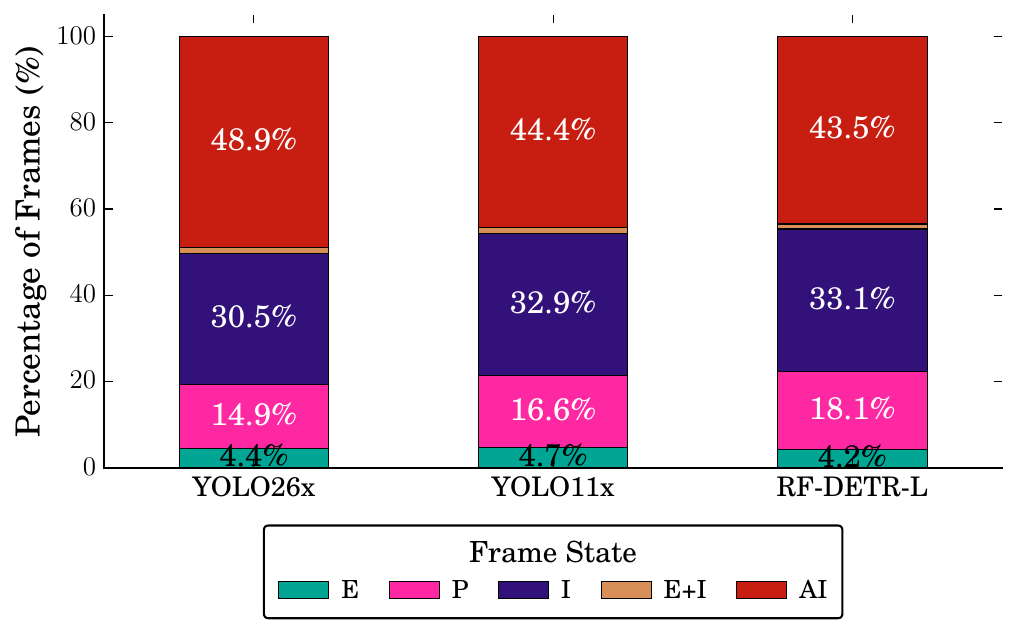}
\vspace{-0.5cm}
\hrule
\vspace{0cm}
\caption{Frame-state distribution of \Albireo{} across three detectors on the BDD100K MOT validation split (200 clips). Skip frames (E + P) account for $19$--$22\%$; the augmented-inference rate (AI) of $43$--$49\%$ measures the rescue mechanism's contribution.}
\label{fig:framestate}
\end{figure}

The five-state taxonomy from \Cref{sec:framestates} classifies each frame as follows: ERD-empty (E), Kalman-predict (P), inference (I), ERD-triggered inference (E+I), and augmented inference (AI, where at least one confirmed object state is rescued by its Kalman prediction). \Cref{fig:framestate} shows the distribution across the three detectors on the BDD100K MOT.

The frame-state breakdown highlights two important behaviors. 
First, the total skip rate $E+P$ is $19.3\%$ for YOLO26x, $21.3\%$ for YOLO11x, and $22.2\%$ for RF-DETR. These are the frames where the full detector is gated, and energy savings accrue. 
Second, rescue is invoked frequently. 
For YOLO26x, AI frames account for $48.9\%$ of all frames, and $61\%$ of frames where the detector runs. 
For YOLO11x and RF-DETR, AI frames account for $44.4\%$ and $43.5\%$ of all frames, respectively, and $56\%$ of detector-invocation frames for both detectors. 
Thus, more than half of the detector invocations leave at least one confirmed object state unmatched, which rescue recovers using the Kalman prediction.  If we did not run 
rescue, those object states would either be removed outright, as in the Kill-only ablation of \Cref{tab:ablation}, or fragment and re-spawn after promotion, creating false negatives over several frames and reducing AP@50. 
The rescue mechanism is a key reason that \Albireo{} saves $14$--$18\%$ energy, while preserving AP@50 within $\pm1.2$\,pp of Vanilla (\Cref{tab:cross_detector}). 
The ERD-triggered inference rate $E+I$ is small ($\sim$1.3\%), reflecting the limited number of empty-to-occupied transitions in the BDD100K MOT, which tends to be object-dense.

\subsection{Ablation: Component Contribution}
\label{sec:eval_ablation}

\Cref{tab:ablation} isolates the contribution of \Albireo{}'s main components on BDD100K MOT with YOLO26x on Thor. 
\emph{Kill-only} removes rescue and deletes confirmed object states after an unmatched detector frame. 
\emph{+ERD} adds the BDD100K-fine-tuned empty-scene screen. 
\emph{+10D KF} replaces the 8D Kalman state with the 10D state used by \Albireo{}, adding size velocity and position acceleration. 
The full \Albireo{} system adds detector-flicker rescue. 
The no-rescue variants cluster between $0.572$ and $0.578$ AP@50, while rescue increases AP@50 to $0.617$. 
Thus, among these components, rescue is the largest contributor to AP@50, while ERD and the 10D state mainly affect skip behavior, high-IoU mAP, and runtime.

\begin{table}[t]
\centering
\caption{Component ablation on BDD100K MOT using YOLO26x on Thor. Values are means over 200 clips.}
\label{tab:ablation}
\footnotesize
\setlength{\tabcolsep}{4pt}
\begin{tabular}{lllcccc}
\hline
\textbf{Variant} & \textbf{KF} & \textbf{Lifecycle} & \textbf{AP@50} & \textbf{mAP} & \textbf{Skip} & \textbf{ms/fr} \\
 & & & & & \textbf{(\%)} & \\
\hline
Kill-only           & 8D CA  & kill         & 0.578 & 0.358 & 21.5 & 41.8 \\
+ERD                & 8D CA  & kill + ERD   & 0.572 & 0.355 & 25.8 & 42.9 \\
+10D KF             & 10D CA & kill + ERD   & 0.577 & 0.407 & 26.3 & 42.5 \\
\textbf{\Albireo{}} & 10D CA & rescue + ERD & \textbf{0.617} & \textbf{0.433} & 19.3 & 46.4 \\
\hline
\end{tabular}
\end{table}

\subsection{Per-Domain Energy and Workload Shift}
\label{sec:eval_perdomain}

\begin{figure}[t]
\centering
\includegraphics[width=\linewidth]{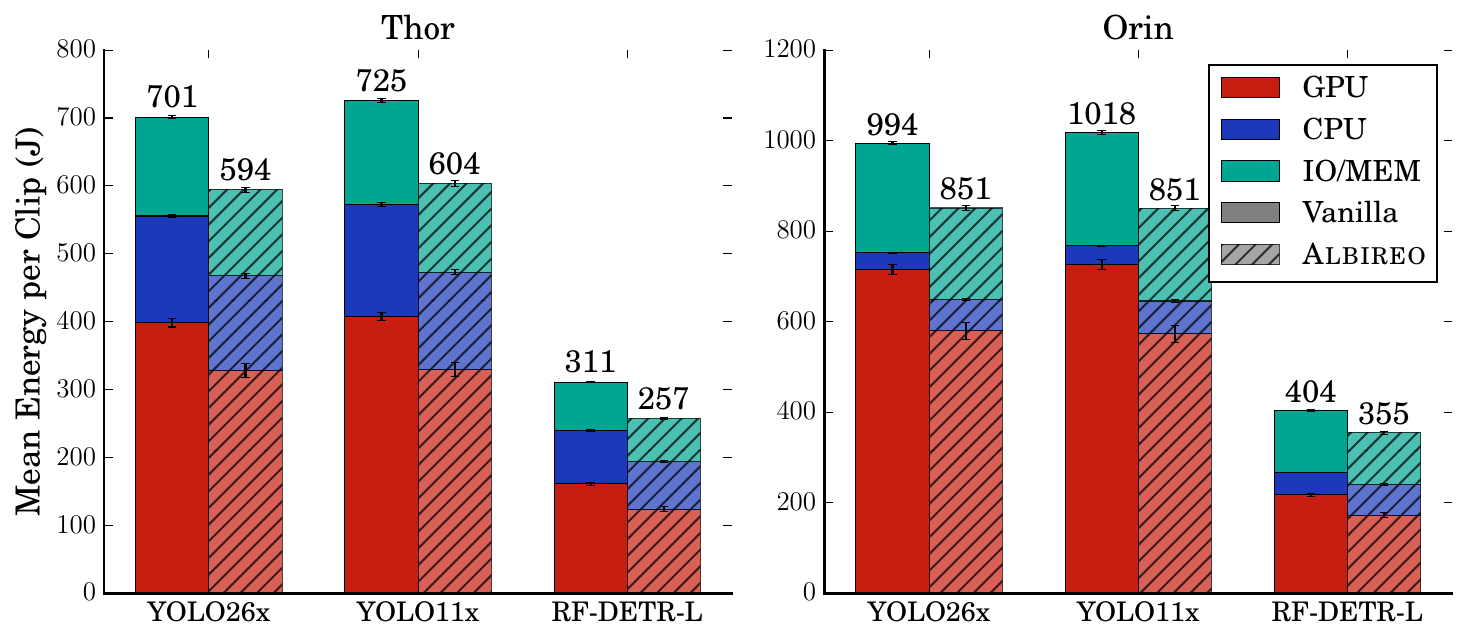}
\vspace{-0.5cm}
\hrule
\vspace{0cm}
\caption{Mapped per-rail energy per clip, grouped as GPU, CPU, and IO/MEM, for Vanilla and \Albireo{} across three detectors on Thor and Orin. IO/MEM denotes memory-related and non-GPU/non-CPU I/O rails exposed by platform telemetry. Grouped rails are used for trend analysis and do not necessarily sum to the top-level board energy reported in \Cref{tab:main}.}
\label{fig:per_domain_energy}
\end{figure}

\Cref{fig:per_domain_energy} breaks per-clip energy into GPU, CPU, and IO/MEM groups using the per-rail power mapping described in~\Cref{sec:eval_setup}. 
The grouped rails are used to explain where the savings come from; total energy is still computed from the top-level board input rail.

Under YOLO26x per-frame inference, GPU power dominates on both platforms. Thor draws $70$\,W total board power, of which $34$\,W ($49\%$) is on the GPU rail; Orin draws $43$\,W total, of which $31$\,W ($72\%$) is GPU. 
Both totals are below the corresponding TDPs, so the workload is not power-throttled. 
\Albireo{} reduces energy primarily by gating GPU activity: for YOLO26x, GPU energy drops by $17.5\%$ on Thor and $19.0\%$ on Orin, exceeding the total-energy reduction. 
CPU and IO/MEM changes are smaller and platform-dependent.

The CPU group shows a platform-dependent cost of control logic. 
On Orin, average CPU power roughly doubles under \Albireo{} from $1.6$\,W to $3.6$\,W, while CPU utilization rises from $\sim$7\% to $\sim$23\%. 
This increase comes from \Albireo{}'s per-frame bookkeeping: Kalman prediction, two-stage IoU/Mahalanobis matching, and rescue accounting run on the CPU, regardless of whether detector inference is triggered. 
On Orin's twelve A78AE cores, this overhead is measurable. On Thor's larger fourteen-core Neoverse V3AE CPU complex, the same operations do not increase CPU power measurably; average CPU power changes from $13.6$\,W under Vanilla to $13.0$\,W under \Albireo{}. 
In both cases, GPU savings dominate and total energy drops, but the CPU shift is important for deployment: adaptive frame-skipping systems that run per-frame bookkeeping on the CPU can trade reduced GPU work for additional CPU-side work.

\subsection{Per-Clip Behavior}
\label{sec:eval_perclip}

\begin{figure*}[t]
\centering
\includegraphics[width=\linewidth]{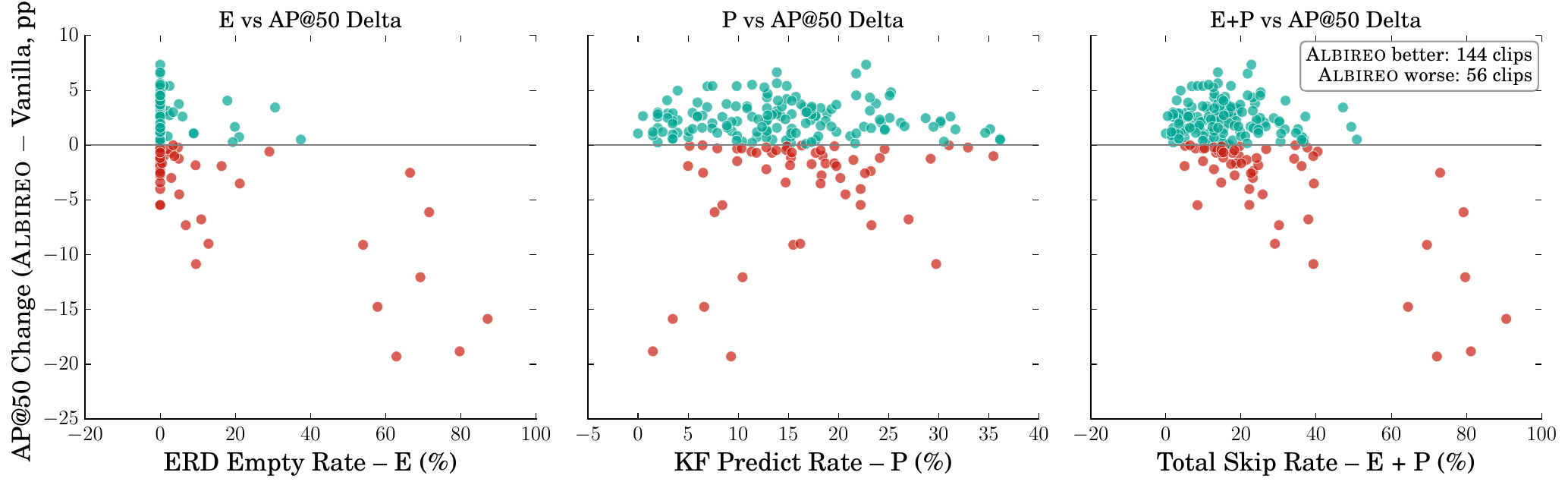}
\vspace{-0.5cm}
\hrule
\vspace{0cm}
\caption{Per-clip skip-rate components versus AP@50 delta (Vanilla $\rightarrow$ \Albireo{}). Across the full P-frame range, AP@50 deltas cluster around zero or positive; large negative deltas concentrate on high-E clips where the absolute ground-truth count is low and AP is more sensitive to individual missed detections.}
\label{fig:scatter_ap_delta}
\end{figure*}

The per-clip view shows that \Albireo{} adapts to scene content rather than imposing a uniform skip cadence. 
Across all 200 clips on Thor with YOLO26x, \Albireo{} achieves the same or better AP@50 as compared to Vanilla on $144$ clips ($72\%$); on the remaining $56$ clips, the median AP@50 drop is below $2$\,pp absolute. 
\Cref{fig:scatter_ap_delta} plots per-clip AP@50 delta against skip rate, decomposed into ERD-empty (E), Kalman-predict (P), and combined skip (E+P) components.

 Using the ego-motion benchmark, the decomposition shows that Kalman-predict skipping (P-frames) is not associated with a systematic AP@50 drop: across the full $0$--$36\%$ P range, per-clip AP@50 deltas remain centered near zero with a narrow spread. 
The clips with the largest negative AP deltas concentrate on the high-E branch ($>40\%$ E), where scenes are mostly empty and the absolute number of ground-truth objects is small. 
In this regime, even a one-frame delay between an ERD non-empty verdict and detector recovery can shift AP@50 by several points. 
Thus, \Albireo{}'s variance is highest on sparse clips, while prediction-only skipping remains stable across dense and moderately occupied clips.

\begin{figure}[t]
\centering
\includegraphics[width=\linewidth]{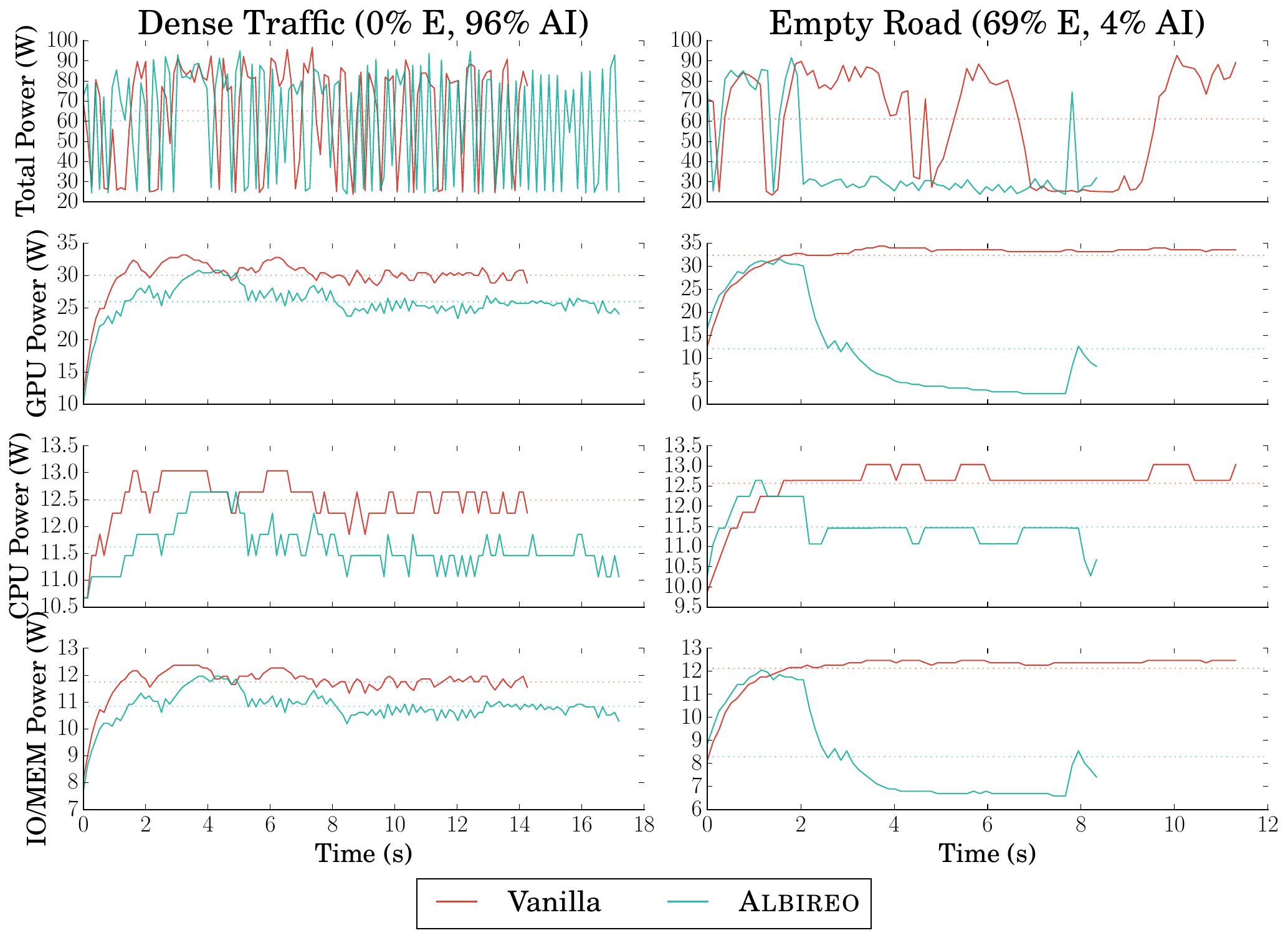}
\vspace{-0.3cm}
\hrule
\vspace{0cm}
\caption{Per-clip power traces (Total, GPU, CPU, IO/MEM) for two representative BDD100K clips on Thor with YOLO26x. Left: dense urban traffic ($0\%$ E, $96\%$ AI), where detector invocations dominate, and rescue preserves AP@50. Right: sparse highway ($69\%$ E, $4\%$ AI), where long ERD-only stretches sharply reduce GPU power.}
\label{fig:power_trace_dual}
\end{figure}

\Cref{fig:power_trace_dual} highlights the two operating modes on representative clips. 
On a dense urban clip (no empty frames; $96\%$ AI rate), \Albireo{} and Vanilla draw similar power because the detector runs on most frames; rescue primarily preserves AP@50 rather than reducing energy. 
On a sparse highway clip ($69\%$ E, $4\%$ AI), \Albireo{}'s GPU power drops sharply during long ERD-only stretches, while Vanilla continues running full detector inference. 
The same algorithm, therefore, produces different power profiles depending on scene content: dense clips emphasize rescue and accuracy preservation, while sparse clips expose the largest energy savings through empty-scene skipping.

\section{Discussion}
\label{sec:discussion}

\subsection{Comparison with the Closest Prior Work}
\label{sec:disc_closest}

The two closest baselines are Statues~\cite{kim2024statues}, which targets energy-efficient video object detection on edge hardware, and CTD~\cite{ding2024ctd}, which uses a Kalman-based skip trigger. \Albireo{} differs from both in the level at which it makes skip decisions and in the deployment assumptions it requires.

\smallskip
\textbf{Versus Statues.} Statues makes pixel-level skip decisions by comparing consecutive frames, which is effective for fixed-camera surveillance, but struggles with ego-motion. In the BDD100K dashcam video, camera motion causes large frame differences even when the object-level scene is stable, so Statues skips only $3.6$--$15.5\%$ of frames across our evaluated thresholds. \Albireo{} instead reasons over object states, so its skip trigger does not rely on a fixed-camera background assumption. On skipped frames, Statues also reuses stale detections, whereas \Albireo{} advances object boxes through the Kalman dynamics, reducing the number of stale-box errors during prediction-only stretches. Finally, Statues was evaluated on custom ASIC/FPGA hardware, while \Albireo{} is a software-only wrapper evaluated on commercial Jetson AGX Orin and Thor platforms.

\smallskip
\textbf{Versus CTD.} CTD is closest to \Albireo{} mechanistically because both use Kalman-filter state to avoid unnecessary detector calls. The key difference is the skip signal. CTD compares the current prediction against the most recent detection using a Mahalanobis-distance score; as consecutive skips accumulate, that reference detection becomes stale. \Albireo{} instead uses covariance growth from the current object state, so the trigger directly reflects how long the state has evolved without measurement. Empirically, at a matched skip rate of $\sim$40\%, \Albireo{} reaches $\mathrm{AP@50}=0.564$ while CTD reaches $0.506$ (\Cref{sec:eval_baselines}). \Albireo{} also adds detector-flicker rescue and conditional empty-scene screening, and reports measured energy savings for this per-object uncertainty-driven policy on commercial edge SoCs.

\subsection{Deployment Implications of CPU-Side Bookkeeping}
\label{sec:disc_workload}

\Cref{sec:eval_perdomain} shows that \Albireo{} trades reduced GPU inference work for additional CPU-side bookkeeping. 
This shift is expected for adaptive frame-skipping systems that run per-frame prediction, association, and rescue logic on the CPU. The detector is invoked less often, but the control logic still runs on every frame. 
On Orin, this overhead is visible in CPU power. On Thor's larger CPU complex, it does not measurably increase CPU power. 
The implication is practical rather than problematic. 
Total energy still drops on both evaluated platforms because GPU savings dominate, but integrators targeting smaller SoCs, such as Jetson Orin Nano or Xavier NX, should account for CPU bookkeeping cost when choosing the uncertainty threshold and rescue policy.

Memory overhead is also limited: across 200 clips in
each of the six detector--platform configurations, \Albireo{} adds only
$2$--$27$\,MB of system memory relative to per-frame inference---below
$0.7\%$ in every case.

\subsection{Composability with Model Compression}
\label{sec:disc_compress}

\Albireo{} reduces the \emph{number} of detector invocations, while model-compression techniques such as quantization, pruning, and knowledge distillation reduce the \emph{cost} of each invocation. 
These two directions are complementary: compression makes detector frames cheaper, and \Albireo{} reduces how often those frames are executed; note that precision choices carry system-level power and energy effects of their own on these SoCs~\cite{taherin2026hydra}. 
If a compressed detector reduces per-inference energy by a fraction $y$ and \Albireo{} skips a fraction $x$ of detector invocations, then the idealized combined reduction is approximately $1-(1-x)(1-y)$, assuming the two effects do not overlap. 
We do not evaluate this combination in the present submission because deploying calibrated compressed versions of all three detectors would require a separate accuracy--energy study. 
However, the mechanisms act on different parts of the pipeline, so compression is a natural extension rather than a competing alternative.

\subsection{Limitations and Future Work}
\label{sec:disc_limits}

Three limitations of the current design point to natural extensions. 
\emph{First}, the constant-acceleration Kalman dynamics fit many vehicle-dominated dashcam scenes, but can drift under highly erratic motion, such as a pedestrian making an abrupt lateral move at a crosswalk. 
The uncertainty trigger is designed to call the detector as uncertainty grows, but the prediction-only box at the moment of abrupt motion can still be inaccurate. 
Adding a learned residual or class-conditioned motion model on top of the Kalman prediction is a natural extension.

\emph{Second}, the empty-scene screen is domain-dependent. 
In this paper, ERD is fine-tuned for BDD100K dashcam scenes, which is appropriate for our evaluation, but not automatically transferable to other camera domains such as indoor surveillance, aerial video, or robot-mounted cameras. 
Future deployments should either fine-tune the lightweight empty-scene classifier for the target domain or replace it with a more general low-cost emptiness oracle.

\emph{Third}, the rescue confidence threshold $c_{\mathrm{rescue}}$ and consecutive-rescue cap $R_{\max}$ are fixed heuristics. 
They work well in our evaluation, but different object classes, object sizes, and motion characteristics may benefit from different rescue lifetimes. 
A learned or adaptive decay policy conditioned on class, confidence history, box size, and motion uncertainty could further improve the AP@50--energy tradeoff.

Finally, our experiments focus on the BDD100K ego-motion video because it is a challenging test case for frame-skipping methods that rely on pixel-level similarity. 
Evaluating \Albireo{} on fixed-camera and non-driving datasets would further characterize how much additional benefit the same object-state trigger provides in easier or structurally different video streams. Our evaluation also isolates a single detection stream: using concurrent execution alongside other workloads on the same SoC was
not evaluated, and co-tenancy evaluation is future work.

\section{Related Work}
\label{sec:related_work}

\Cref{tab:rw_comparison} in \Cref{sec:gap} summarizes the deployment properties most relevant to \Albireo{}; here we discuss the broader literature in more detail.

\subsection{Temporal Feature Propagation}
\label{sec:rw_temporal}

A large body of video object detection work exploits temporal redundancy by propagating information across frames. 
Deep Feature Flow (DFF)~\cite{zhu2017dff}, Flow-Guided Feature Aggregation (FGFA)~\cite{zhu2017fgfa}, the unified framework of Zhu et al.~\cite{zhu2018high}, and YOLOV~\cite{shi2024yolov} propagate deep feature maps from sparse keyframes using optical flow or learned aggregation. 
Other methods aggregate temporal context through spatial-temporal memory~\cite{xiao2018stmn}, sequence-level semantic aggregation~\cite{wu2019selsa}, tubelet proposals and tubelet rescoring~\cite{kang2017tubelet,kang2017tcnn}, recurrent-convolutional feature propagation~\cite{liu2018mobile}, or scale-time lattices~\cite{chen2018scaletime}. 
These methods improve the speed--accuracy tradeoff for video detection, but they typically require detector-specific temporal modules, architecture changes, or end-to-end retraining.

Some temporal methods improve detection quality without reducing detector invocations. 
Seq-NMS~\cite{han2016seqnms}, for example, links detections across frames as a post-processing step and boosts weak detections using temporal consistency. 
Because the detector still runs on every frame, such methods do not directly address the edge-energy cost targeted by \Albireo{}. 
\Albireo{} instead leaves the detector architecture unchanged and controls when detector inference is invoked.

\subsection{Frame Skipping and Adaptive Inference}
\label{sec:rw_frame_skipping}

Frame-skipping systems reduce cost by avoiding detector execution on selected frames. 
SDOF-Tracker~\cite{nishimura2022sdof} skips human detection at a fixed interval and fills intermediate frames using optical-flow-based tracking. 
Park and Kim~\cite{park2020frameskip} propose frame skipping for fast object detection, and Ahmed et al.~\cite{ahmed2021drone} apply dynamic frame skipping to drone-based multi-object detection. 
These methods share \Albireo{}'s goal of reducing redundant detector calls, but fixed or heuristic schedules remain content-blind and can skip frames where fresh detections are needed.

Other approaches adapt the amount or location of computation. 
Ohashi and Yokoyama~\cite{ohashi2025omnidirectional} use background subtraction to crop moving regions in ultra-high-resolution omnidirectional video. 
Skip-Convolutions~\cite{habibian2021skipconv} skips redundant convolution operations across frames, while PASS~\cite{zhou2024pass} skips temporally unchanged patches for on-device video perception and evaluates the idea on Jetson Nano. 
These methods reduce computation below the frame level, but require model modification and patch-level routing. 

Learned scheduling methods decide when or how to run detection using trained policies. 
Luo et al.~\cite{luo2019detect} train a scheduler to choose between detection and tracking, Arefeen et al.~\cite{arefeen2022framehopper} use reinforcement learning to learn skip-length policies, and Xu et al.~\cite{xu2022litereconfig} adapt model variants and resolutions at runtime on Jetson TX2/Xavier. 
These systems are effective, but their policies require offline training or tuning tied to a detector, dataset, or deployment setting. 
\Albireo{} instead uses object-state uncertainty as a runtime control signal and does not modify or retrain the detector.

Several systems measure or target energy directly on edge platforms. 
Henning et al.~\cite{henning2023framedrop} study frame dropping for multi-object tracking and report up to $28\%$ energy savings on KITTI with a $6.6\%$ Higher Order Tracking Accuracy (HOTA) drop. 
Contoli et al.~\cite{contoli2024dipm} adapt the inference frame rate based on scene dynamics on Jetson Nano, achieving a $21$--$36\%$ energy reduction. 
These studies validate the energy benefits of adaptive video inference, but use a fixed or heuristic frame-rate adjustment rather than a per-object uncertainty-driven detector gating.

Streaming perception~\cite{li2020streaming} studies latency-aware
scheduling and forecasting under asynchronous frame arrivals.
\Albireo{} addresses the complementary question of whether the detector
should be invoked at all, reducing average per-frame latency and energy.

\subsection{Compressed-Domain Video Inference}
\label{sec:rw_compressed}

Compressed-domain methods reduce video inference cost by reusing codec information. 
Wang et al.~\cite{wang2019compressed} run full detection on I-frames and propagate detections to P/B-frames using motion vectors and residuals. 
Tran et al.~\cite{tran2023fast} similarly run a full detector on I-frames and propagate detections to P-frames using lightweight Shift and Refine networks trained with compressed-domain motion vectors, reporting $3$--$20\times$ speedups on high-resolution video. 
True and Khan~\cite{true2023motion} extrapolate bounding boxes between keyframes using video-bitstream motion vectors, and Liu et al.~\cite{liu2022compressed_mot} apply a similar key/non-key-frame split to multi-object tracking with a lightweight tracking CNN. 
MA-YOLO~\cite{wang2025mayolo} propagates YOLO detections using H.264 motion vectors and XGBoost-based offset regression. 
More recent methods, such as Duch\'{e} et al.~\cite{duche2026compressed} and ComPrivDet~\cite{yao2026comprivdet}, further exploit compressed-domain cues: Duch\'{e} et al. report up to $3.7\times$ speedup with only $4\%$ mAP@0.5 drop, while ComPrivDet skips over $80\%$ of inferences for privacy-object detection.

These systems share \Albireo{}'s goal of reducing full detector invocations, but they depend on codec metadata such as H.264/H.265 motion vectors and residuals. 
That assumption is restrictive for raw camera pipelines, including direct sensor feeds on embedded platforms before video encoding. 
Many compressed-domain systems also follow codec-imposed keyframe schedules rather than deciding from per-object uncertainty when detection is needed. 
\Albireo{} operates on decoded frames and object-state covariance, requiring no codec infrastructure or fixed group-of-pictures (GOP) schedule.

\subsection{Model Compression and Tracking-Based Detection}
\label{sec:rw_compress_tracking}

Model compression techniques such as quantization, pruning, and knowledge distillation reduce the cost of each detector invocation. 
For example, lightweight YOLO variants have been proposed for efficient vehicle detection~\cite{tajar2021lightweight}. 
These techniques are complementary to \Albireo{}: compression reduces the cost of each detector call, while \Albireo{} reduces the number of calls. 
Running a compressed detector inside \Albireo{} could therefore further increase savings.

Tracking-based detection also exploits temporal structure. 
Feichtenhofer et al.~\cite{feichtenhofer2017detect} jointly train detection and tracking components so that temporal correlation improves both tasks. 
Liu et al.~\cite{liu2023objects} predict future object locations from a single frame to maintain detections through temporary misses. 
These methods share \Albireo{}'s intuition that object locations are temporally predictable, but they require trained temporal modules or joint detector modification. 
\Albireo{} instead uses a lightweight Kalman object state  outside the detector to decide when to invoke inference and when to emit prediction-backed boxes after brief detector misses.

\section{Conclusion}
\label{sec:conclusion}

In this paper, we have presented \Albireo{}, a detector-agnostic, codec-free, content-aware frame-skipping system for edge video object detection that requires no detector modification or detector retraining. 
\Albireo{} combines three mechanisms: a forward-looking uncertainty trigger that gates detector invocation using per-object Kalman covariance growth, a detector-flicker rescue mechanism that preserves confirmed object states through brief detector misses, and a lightweight empty-scene screen that avoids full detector calls on objectless frames. 
Across BDD100K MOT, three detector architectures, and two Jetson SoC generations, \Albireo{} keeps AP@50 within $1.2$\,pp of per-frame inference while reducing energy by $12.1$--$17.6\%$. 
On the primary YOLO26x configuration, the default operating point Pareto-dominates per-frame inference on AP@50, energy, and latency, and the uncertainty-threshold sweep forms the upper Pareto frontier over FixedSkip-$N$, CTD, and Statues in the target operating region ($\mathrm{AP@50}\geq0.50$). 
These results show that using the object-level temporal state can serve as a practical drop-in efficiency layer for off-the-shelf edge video detection pipelines.

\section*{Acknowledgment}
We thank the anonymous reviewers for their constructive
feedback. This work was partially supported by the U.S. National Science
Foundation (NSF) SaTC program under Grant No.~2414652, the EU Project
dAIEDGE (GA Nr 101120726), and the Innovate UK Horizon Europe Guarantee (GA Nr
10090788). In preparing this paper, the authors used generative AI tools
(OpenAI ChatGPT and Anthropic Claude) for language and presentation
refinement. All technical content, results, and claims were produced and
verified by the authors, who take full responsibility for the content of
this paper.

\bibliographystyle{IEEEtran}
\balance
\bibliography{references} 

\end{document}